\documentclass[10pt,twocolumn,letterpaper]{article}

\usepackage{iccv}
\usepackage{times}
\usepackage{epsfig}
\usepackage{graphicx}
\usepackage{amsmath}
\usepackage{amssymb}

\usepackage{color, colortbl}
\definecolor{Gray}{gray}{0.9}
\usepackage{subfigure}

\usepackage[pagebackref=true,breaklinks=true,letterpaper=true,colorlinks,bookmarks=false]{hyperref}

\iccvfinalcopy 

\def\iccvPaperID{9576} 

\ificcvfinal\fi

\begin{document}

\title{Fine-Grained Re-Identification}

\author{Priyank Pathak\\
New York University\\
{\tt\small ppriyank@nyu.edu}
}

\maketitle
\ificcvfinal\thispagestyle{empty}\fi

\begin{abstract}
  Research into the task of re-identification (ReID) is picking up momentum in computer vision for its many use cases and zero-shot learning nature. This paper proposes a computationally efficient fine-grained ReID model, FGReID, which is among the first models to unify image and video ReID while keeping the number of training parameters minimal. FGReID takes advantage of video-based pre-training, spatial attention, and non-local operations to excel on both video and image ReID tasks. FGReID achieves state-of-the-art (SOTA) on MARS, iLIDS-VID, and PRID-2011 video person ReID benchmarks. Eliminating temporal pooling yields an image ReID model that surpasses SOTA on CUHK01 and Market1501 image person ReID benchmarks. The FGReID achieves near SOTA performance on the vehicle ReID dataset VeRi, demonstrating its ability to generalize. Additionally, we do an ablation study analyzing the key features influencing model performance on ReID tasks. Finally, we discuss the moral dilemmas related to ReID tasks, including the potential for misuse. Code for this work is publicly available at  
  \url{https://github.com/ppriyank/Fine-grained-Re-Identification}.

\end{abstract}

\section{Introduction}

Re-identification (ReID) aims to match identical objects such as people (person ReID), vehicles (vehicle ReID) and faces (face ReID), that experience subtle variations across camera viewpoints and time. ReID is most useful for identifying unique objects for surveillance purposes. However, ReID can also be used to create embeddings for generic search engines. While image ReID depends on a single image for ReID, it is susceptible to complexities such as occlusion, low illumination, and inferior viewpoints. Video ReID uses multiple frames across time to counteract any identification impediments in a single image. The downside of video ReID over image ReID is the added complexity of handling additional temporal information.

\begin{figure}[!t]
 \centering
  \includegraphics[width=1\linewidth]{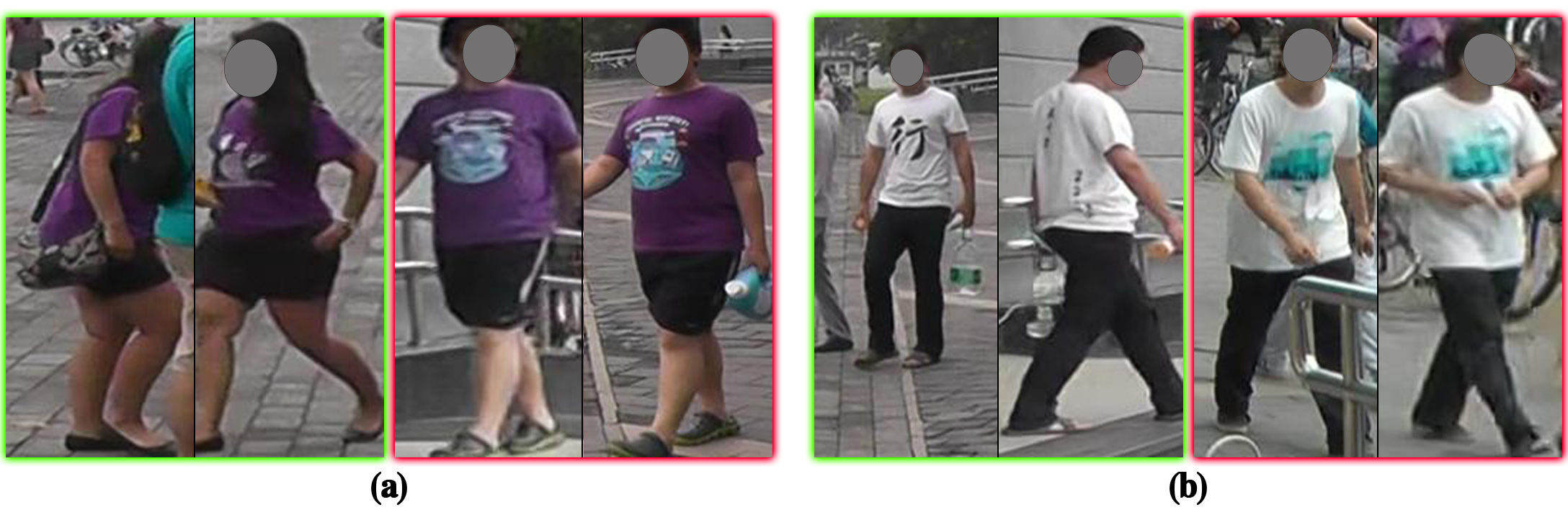}
 \caption{(a) and (b) show two examples of misclassifications from the MARS dataset with similar clothing styles made by a spatial averaging model \cite{pathak2019video}.}
 \label{fig:mars_misclass}
 \end{figure}

Existing person ReID works have utilized graph convolutional neural networks \cite{graph5, graph4, graph3}, self-attention  \cite{ilvid3, transformer2}, temporal attention \cite{attn_better_lstm, Tattn2, Tattn1}, and LSTMs \cite{lstm1}. Many of these works depend on the object's appearance (spatial structure), overlooking the fine-grained subtleties containing important distinguishing attributes. For example, existing approaches often fail when people have similar clothing styles, as demonstrated in Figure \ref{fig:mars_misclass}. The aforementioned methods are also computationally expensive and impose restrictions on the input image size and batch size, critical to performance improvement. Numerous image ReID models cannot generalized to videos and vice versa, discarding surplus hours of available surveillance footage. We propose a unified approach for video and image ReID that requires fewer training parameters and uses fine-grained details to re-identify individuals more accurately.

 The task of learning fine-grained details can be divided into implicit and explicit methodologies. Implicit methods include parts-based matching \cite{str_reid1, veri1, str_reid4}, which are often contingent on the alignment of parts, and feature erasing \cite{global_narrow} to learn minute details indirectly. Explicit methods \cite{fine_erase1, fine_grain_erase2} commonly deploy spatial attention maps to isolate and highlight subtle details. While such methodologies perform very well for images, they are not scalable for long videos. In this work, we extend an explicit fine-grained image classification technique \cite{finegrained} to videos by employing an extra ResNet network for only handing minute details. Such an approach is computationally cheap compared to other fine-grained ReID models, which involve multiple passes.

 \begin{figure}[!t]
 \centering
 \includegraphics[width=1\linewidth]{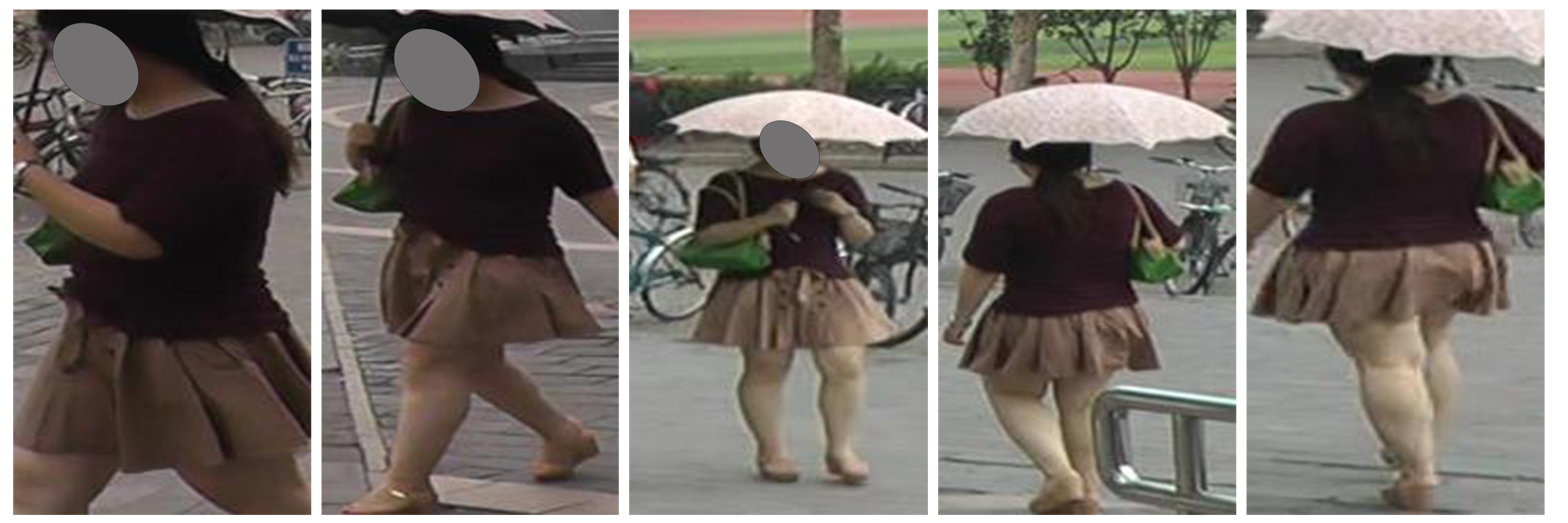}
 \caption{Video frames from the MARS dataset. Although there is some occlusion by the umbrella, both front and back views and a distinctive green purse are clearly visible.}
 \label{fig:attn_network2}
 \end{figure}
 
\begin{figure}[!t]
 \centering
 \includegraphics[width=1\linewidth]{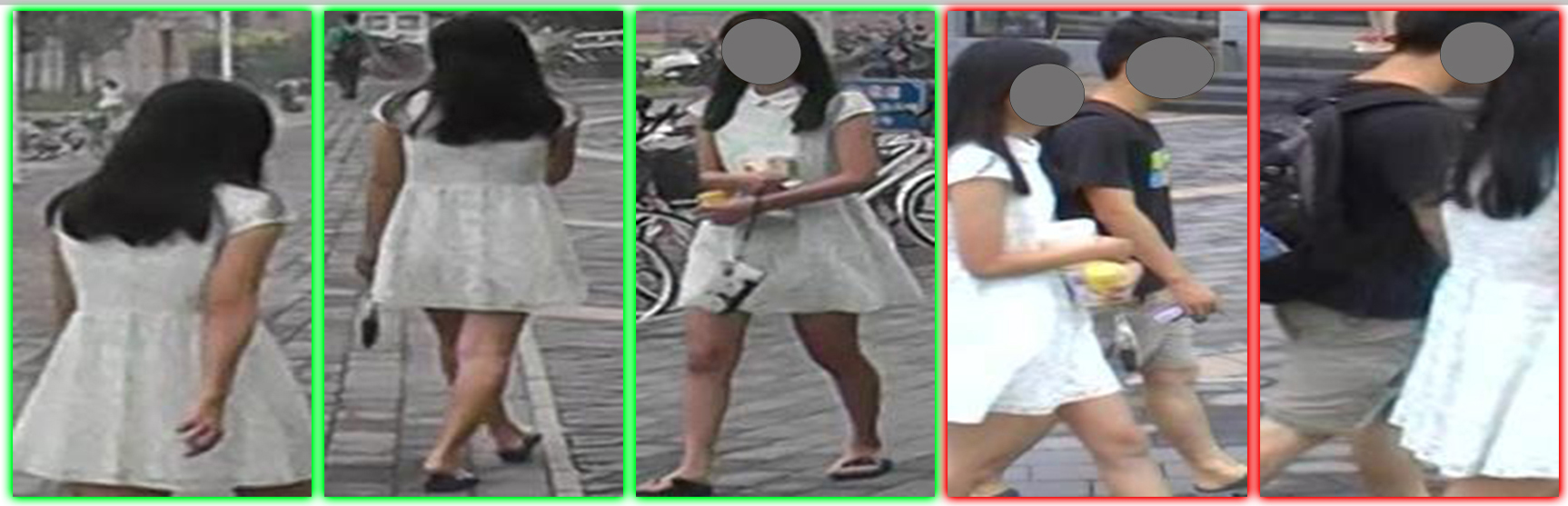}
 \caption{MARS dataset video frames of the same individual. Within the same video, green-highlighted and red-highlighted frames focus on different individuals.}
 \label{fig:attn_network1}
 \end{figure}
 
Video person ReID models should be able to handle both temporal and spatial variations. Numerous works \cite{attn_better_lstm, Tattn2, Tattn1} use a specialized temporal attention model to merge temporal information. Such models often assign smaller weights to frames with heavy occlusion, but that may cause loss of distinctive features.  Fu \etal \cite{str_reid3} argue that occluded frames still contribute characteristic details like those in Figure \ref{fig:attn_network2}. Su \etal \cite{context_frame1} demonstrated some of the non-occluded outlier frames might confuse a model in targeting the wrong individuals. Figure \ref{fig:attn_network1} shows the red highlighted frames focusing on the black-dressed individual as opposed to the correct green highlighted frames. In such cases, temporal attention fails, favoring computationally cheap temporal pooling. We also include non-local operations \cite{nonlocalattn} to introduce context that can disregard outlier frames. 

Our contributions are as follows. 1- We adopt a fine-grained image classification model and propose a novel framework, FGReID, capable of generating contextually-aware fine-grained embeddings for images and videos. It's ability to re-identify is not only limited to people but vehicles as well. FGReID is among a few frameworks that perform equally well on images and videos, making it optimal for a wide range of ReID problems. 2- The model employs a limited number of training parameters making it computationally inexpensive and lightweight while capturing fine-grained details in one pass. We show FGReID size and computation time comparison with publicly available video ReID models. 3- Extensive experiments show FGReID exceeding SOTA on two large-scale video person ReID datasets MARS and iLIDS-VID, while matching SOTA on the PRID-2011 dataset. For image ReID, our model exceeds SOTA on Market1501, CUHK01, while it is on par with SOTA on the VeRi-776 dataset. 4- We also address the ethical concerns regarding ReID work.

\section{Related Work}

\begin{figure*}[!ht]
 \centering
 \includegraphics[width=1\linewidth]{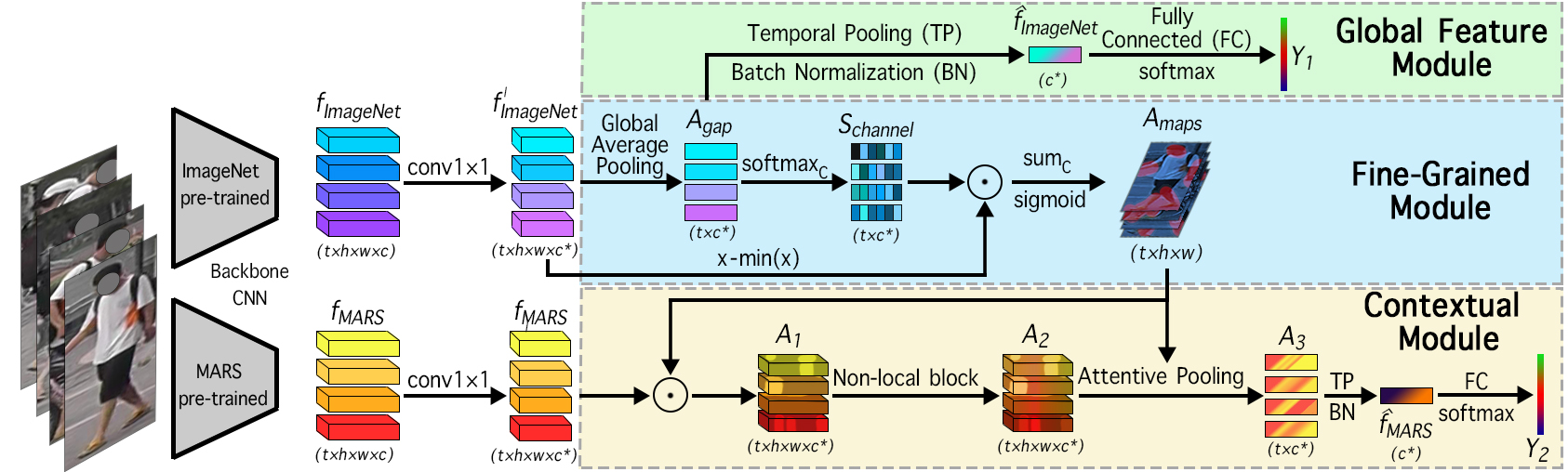}
 \caption{The proposed video model has shown $t = 4$ video frames as input. Backbone CNN creates preliminary features, followed by a dimension reduction step. Global Feature Module creates a general overview of the entire clip. The fine-grained module highlights the spatial intricacies while the context module adds context via a non-local block. $\odot$ represent element-wise multiplication, $softmax_c$, and $sum_c$ represent softmax, and summation along the channel dimension.}
 \label{fig:video_architecture}
 \end{figure*}
 
\noindent \textbf{Image ReID}: Recent image person ReID work either relies on deep learning models \cite{transformer2, transformer1, cuhk3} or on techniques \cite{variance_reg, bot, market1, market2, reranking, market3} for boosting the performance of existing frameworks. Recently, Liu \etal \cite{market5} utilized generative adversarial networks (GAN) to produce style invariant images from various camera viewpoints. GANs often blur pictures in recreating scenes that might miss crucial distinctive features. Many image ReID models mistake similar-looking entities for each other, especially in Vehicle ReID. 

Vehicle ReID is conceptually identical to person ReID, aiming to retrieve vehicles rather than people. Cars lack distinctive features and have similar background roads. In such cases, subtle details play a much more vital role in resolving similar vehicles. Prior vehicle ReID work includes handcrafted \cite{vehicle1, vehicle2} and deep learned features  \cite{veri1, Tattn2, vehicle4}. He \etal \cite{fastreid} is among the principal attempts to unify vehicle and person ReID. Image ReID methods are computationally expensive and infeasible for handling the temporal relations, making them unscalable for videos.

\noindent \textbf{Video person ReID}:
Common methodologies for generating video embeddings involve transformers \cite{action_transformer}, 3D convolutions \cite{3dconv, 3dconv4, 3dconv2}, RNNs and LSTMs \cite{lstm1, lstm3, lstm2}. While the results are promising, these approaches often have memory constraints for a single GPU and involve many training parameters, making them infeasible for real-world surveillance systems. RNN based approaches often neglect intricate details within the frames, concentrating more on intra-frame structural ties. Gao and Nevatia \cite{attn_better_lstm} showed a temporal attention model could outperform LSTMs and 3D convolutions for ReID. Many 2D convolution-based methodologies have achieved significant ReID success, specifically those involving attention modules \cite{str_reid3, Tattn1, div_reg, context_frame1}.

Graph-based methods \cite{graph1, str_reid1, graph5, graph4} have been applied to both image and video person ReID. Shen \etal \cite{graph2} treat images as graph nodes while disregarding the spatial subtleties.  Yang \etal \cite{graph3} use two branches for generating spatial and temporal relations, where nodes of graphs are segments of images. These approaches incur heavy computation and are ineffective at differentiating similar clothing styles, where the spatial structure is identical.

\noindent \textbf{Fine-Grained}: Most fine-grained related research deals with images, either through enhancing image quality or recursively cropping critical regions and generating embeddings concurrently \cite{fine_erase1, fine_grained_3, fine_grain_erase2}. Such approaches fail on long videos. 
Generally, fine-grained ReID work has shown moderate success \cite{fine_grained_5, global_narrow,fine_grained_6}. Recently, Zhang \etal \cite{mars_sota} proposed multi-granular attention for videos, surpassing the state-of-the-art on MARS, iLIDS-VID, and PRID-2011 datasets. While such an approach is promising, but it incurs a high computational cost by involving multiple passes of the same image for capturing different granularity or structural details. Hence, we build upon the computationally efficient approach of Zhu \etal \cite{finegrained}, involving a second ResNet-50 as a mechanism for generating fine-grained rich embedding in one-pass. 

\section{Methodology}
\label{section:methodology}
Figure \ref{fig:video_architecture} shows our proposed model architecture FGReID. FGReID has three major segments: a Global Feature Module (Section \ref{section:gloabal_features}), a Fine-Grained Module (Section \ref{section:fine_grained}), and a Context Module (Section \ref{section:context_module}). The Global Feature Module averages feature spatially and temporally producing coarse-grained features ($\hat{f}_{ImageNet}$). The Fine-Grained Module creates spatial attention maps inspired by work on fine-grained image classification \cite{finegrained} in a parameter-less manner. The Context module creates context-aware embeddings ($\hat{f}_{MARS}$) with the help of fine-grained spatial attention and non-local operations. Concatenating $\hat{f}_{ImageNet}$ and $\hat{f}_{MARS}$ produces the final embeddings $f^*$. A shared weight classifier with softmax activation creates label vectors $Y_1$ and $Y_2$ corresponding to $\hat{f}_{ImageNet}$ and $\hat{f}_{MARS}$, respectively. 
The entire architecture employs only five $1\times1$ convolutions and two backbone ResNet-50(s), a single classifier, and two batch norm layers. We shall follow the $t\times h\times w\times c$ convention for the subsequent discussion, indicating a feature having $t$ frames ($t=1$ for images) with $(h, w)$ spatial points and $c$ channels.

\subsection{Backbone CNN Network}
\label{section:backbone_CNN}
We use ResNet-50 as our backbone CNN for generating features for each video frame. We expand the receptive field of ResNet-50 by adjusting the last stride from $(2,2)$ to $(1,1)$ as described by Luo \etal \cite{bot}. Further, we follow the fine-grained image classification approach \cite{finegrained} of utilizing two ResNet-50(s): $CNN_{ImageNet}$ and $CNN_{MARS}$. $CNN_{ImageNet}$ is the generic ImageNet pre-trained CNN trying to capture the coarse-grained features of the image. Simultaneously, the other ResNet-50 ($CNN_{MARS}$) is the ReID task-specific CNN trying to capture fine-grain details obtained by pre-training ResNet on a large person ReID dataset (MARS dataset, ResNet-50 produced from Pathak \etal \cite{pathak2019video}). Both ResNet-50(s) are jointly optimized and trained in an end-to-end fashion. $CNN_{ImageNet}$ and $CNN_{MARS}$ produce $f_{ImageNet}$ and $f_{MARS}$, followed by $1\times1$ convolution for dimensional reduction, producing $f_{ImageNet}^{'}$ and $f_{MARS}^{'}$ respectively (both $\in\mathbb{R}^{t\times h\times w\times c^*}$).

\subsection{Global Feature Module}
\label{section:gloabal_features}
Global Feature Module works on the $CNN_{ImageNet}$, capturing the overall foreground and background of the entire input video/image. The module uses global average pooling to average spatial features. The resulting feature vectors $A_{gap}$ are temporally averaged via temporal pooling (not needed for images). Finally, features are batch-normalized, producing coarse-grained features $\hat{f}_{ImageNet}$. 

\subsection{Fine-Grained Module}
\label{section:fine_grained}
Fine-Grained Module produces fine-grained spatial attention maps, highlighting subtle intricacies within the spatial feature space. Traditionally, a weighted sum of channels learned via 2D convolution highlights these spatial regions. We argue against these pre-trained channel weights, which may not generalize well for unseen test classes. Hence, we deploy a run-time channel weighting technique (ideal for ReID problems) based on the intuition that the significance of a particular channel is proportional to the average activation of its spatial feature map. We use spatially averaged features $A_{gap}$ for weighing the channels.
\begin{align} \label{eq:normalization}
   S_{channel} &= softmax_c(A_{gap}) & \in \mathbb{R}^{t\times c^*}
\end{align}
where $softmax_c$ is the softmax operation along the channel dimension. We apply these channel weights $S_{channel}$ to the absolute value of feature maps with sigmoid activation ($\sigma$):
\begin{align}
   & f_{ImageNet}^{+} = f_{ImageNet}^{'} - min\big(f_{ImageNet}^{'}\big)  \\ 
&    A_{maps} = \sigma \bigg(\sum^{c^*} f_{ImageNet}^{+} \odot S_{channel}\bigg) \hspace{.1cm}  \label{eq:attn_map1}
\end{align}
where $A_{maps} \in \mathbb{R}^{t\times h\times w}$ is the spatial heat map, and $min$ returns the minimum value in the entire tensor. $\odot$ refers to pairwise multiplication of vectors. Such an approach computes spatial attention without adding training parameters. 

\subsection{Context Module}
\label{section:context_module}
\begin{figure}[!t]
 \centering
 \includegraphics[width=1\linewidth]{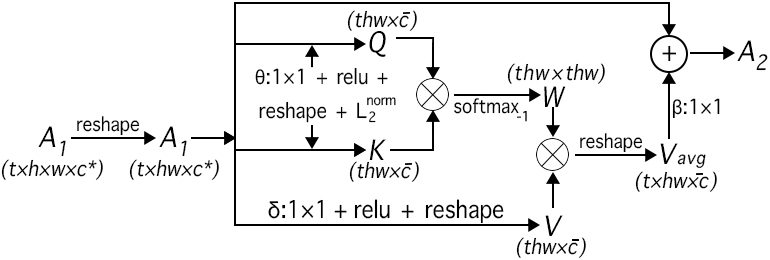}
 \caption{Non-local block pipeline. $1\times1$ convolution is on the $T$ and $HW$ vector space (for images, its the $H$ and $W$ vector space), creating Query (=Key) and Value vectors. $\otimes$ and $\oplus$ indicates tensor multiplication and addition. $softmax_{-1}$ means softmax on each row.}
 \label{fig:nonlocalnn}
 \end{figure}

 We use the context module via a non-local block \cite{nonlocalattn} (self-attention) on the task-specific features $f_{MARS}^{'}$ to introduce spatial and temporal context. The non-local block operation does a weighted sum of each point in spatial and temporal space ($THW$ space) to add context to the features. This approach is susceptible to noisy outliers, which may weaken the weight of significant regions. Hence, as an added safeguard, we shift our non-local block after the spatial attention to reduce the contribution of irrelevant regions.
\begin{align} 
   & A_1 =  f_{MARS}^{'} \odot  A_{maps} \hspace{.5cm} \in \mathbb{R}^{t\times h\times w\times c^*}
\end{align}
We pass the spatially attended features ($A_1$) through the non-local block, as shown in Figure \ref{fig:nonlocalnn}. The traditional non-local block consists of a softmax-ed dot product between the query ($Q$) and key ($K$) vectors to assign weights to each key-value pair, followed by a weighted sum of value ($V$) vectors. The key-value pair act as the context for the query in $THW$ feature space (for images, its $HW$ feature space), giving more weight to contextually similar value vectors. We keep the query and key vectors same, saving around 0.497\% (0.26 million) parameters. 
\begin{table*}[!ht]
\begin{center}
\begin{tabular}{|l|c|c|c|c|c|c|c|c|c|c|}
\hline
 \multicolumn{2}{|c|}{Description} &  \multicolumn{2}{|c|}{Train Set}  & \multicolumn{2}{|c|}{Query Set} & \multicolumn{2}{|c|}{Gallery Set} & \multicolumn{2}{|c|}{Total}  & Split \\ \hline
 Image Dataset & Cameras & IDs  & Images & IDs & Images  & IDs & Images & IDs & Images & \\
\hline\hline
Market1501 \cite{market_d} & 60 & 751 & 12936 & 750 & 3368 & 751 & 19732 & 1501 & 36036 & PD \\ 
CUHK01 (p=486) \cite{cuhk_d} &4 & 485 & 1940 & 486 & 972 & 486 & 972  & 971 & 3884 & AVG:10 \\  
CUHK01 (p=100)  \cite{cuhk_d} & 4& 871 & 3484 & 100 & 200 & 100 & 200  & 971 & 3884 & AVG:10\\   
VeRi-776 \cite{veri_d} & 20 & 576 & 37778 & 200 & 1678 & 200 & 11579  & 776 & 51035 & PD \\
\hline\hline
Video Dataset & Cameras & IDs  & Tracklets & IDs & Tracklets  & IDs & Tracklets  & IDs & Tracklets &  \\
\hline\hline
MARS \cite{mars_d} & 6 & 625 & 8298 & 626 & 1980 & 622 & 9330 & 1251 & 19608 & PD  \\ 
iLIDS-VID \cite{ilvid_d} & 2&150 & 300 & 150 & 150 & 150 & 150 & 300 & 600 & PD \\ 
PRID-2011 \cite{prid_d} & 2&89 & 178 & 89 & 89 & 89 & 89  & 178 & 356 & PD \\ 
\hline
\end{tabular}
\end{center}
\caption{Datasets used for the experiments. Tracklets refer to clips of frames. IDs mean unique identities in each split. Split refers to train and test set division, where PD means predefined split and AVG:10 means an average of ten random splits.}
\label{table:datasets}
\end{table*}
\begin{align}
    Q = K &= L_2^{norm} (reshape(  relu(\theta_{1\times 1} (A_1)))) \\
     V &=  reshape(relu(\delta_{1\times 1} (A_1)))
\end{align} 
where $\theta$ and $\delta$ are $1\times1$ convolution reducing channel dimension to $\overline{c}$(=$\frac{c^*}{4}$). $L_2^{norm}$ does a $L_2$ normalization of $THW$ vectors. Matrix multiplication ($\otimes$) between query $Q$ and key $K$ produces weight matrix $W$
\begin{align} 
    W &= softmax((Q)^T \otimes  K)&\in\mathbb{R}^{thw\times thw} \\ \label{eq:weighingvalue} 
    V_{avg} &= reshape(W \otimes V)&\in\mathbb{R}^{t\times h\times w\times\overline{c}}
\end{align}
We use equation \ref{eq:weighingvalue} to do a weighted sum of each value vector $V$, followed by reshaping and $1\times1$ convolution ($\beta$) to restore input tensor dimension. Adding the resulting vector to the original $A_1$ yields contextually aware feature $A_2$:
\begin{align} 
   A_2 = \beta_{1\times1}(V_{avg}) + A_1  \in  \mathbb{R}^{t\times h\times w\times c^* }
\end{align}
Additionally, we do weighted spatial averaging on $A_2$ via attentive pooling (Equation \ref{eq:attnpooling}):  
\begin{align} \label{eq:attnpooling}
  A_{3}= \frac{\sum^{(h,w)} A_2}{\sum^{(h,w)} A_{maps}} \in \mathbb{R}^{t \times c^* }
\end{align}
Final features $\hat{f}_{MARS}$ is obtained by passing $A_3$ through temporal pooling and batch normalization.

\section{Experiments}
\subsection{Loss Functions} 
We apply labeled smoothed cross-entropy loss exclusively on each label vector $Y_1$ and $Y_2$ and use their mean as the final classification loss $L_{CE}^{avg}$. Among other loss functions, we apply batch hard triplet loss function ($L_{trip}$), center loss ($L_{C}$), and CL Centers OSM loss ($L_{OSM}$)\cite{pathak2019video, osm_caa} on the final features $f^*$. Additionally, we apply a variant of variance regularization \cite{variance_reg}, which penalizes the same class feature $(f^*)^{c^\circ}$ variance across batch $B$, where $c^\circ$ indicates the unique set of identities in $B$. 
\begin{equation} \label{eq:var_reg}
R_{var}=  \sum_{c^\circ \in B}  \Bigg( \frac{1}{ \sum^{c^\circ}_i 1   } \sum^{c^\circ}_i  \Big(  (f^*)^{c^\circ}_i  - \frac{
\sum^{c^\circ}_i (f^*)^{c^\circ}_i}{ \sum^{c^\circ}_i 1 }   \Big)^2    \Bigg)
\end{equation}
where $(f^*)^{c^\circ}_i$ indicates $i^{th}$ instance for the class $c^\circ$. $ \frac{
\sum^{c^\circ}_i (f^*)^{c^\circ}_i}{ \sum^{c^\circ}_i 1 }  $ is the mean of the features belonging to the same class $c^\circ$ within a batch $B$.

Additionally, we apply KL divergence ($L_{cns}$) to align coarse-grained predictions Y1 with the fine-grained ones Y2.
We also use satisfied rank loss $L_{sr}$ \cite{finegrained} to prevent $Y_1$ from dominating $Y_2$, consisting of an unbounded rank loss ($L_{r}$) and a limiting loss ($L_{s}$).
We do hyperparameter optimization \footnote{\url{https://github.com/facebook/Ax}} for MARS and iLIDS-VID datasets for determining weights for each loss, which are used for training and evaluating all the other datasets.  
The total loss comprises a  weighted sum of all the losses mentioned above.
\begin{equation} \label{eq:toal_loss}
\begin{split}
L_{T} & = L_{CE}^{avg} + (1-\beta) * L_{trip} + \beta * L_{OSM} + W^{var} * R_{var} \\
 &  + W^{c} * L_{C} + W^{KL} * L_{cns} + W^{sr} * L_{sr} 
\end{split}
\end{equation}
\subsection{Datasets and Evaluation Metrics}
Table \ref{table:datasets} summarizes various datasets used for our experiments. For evaluation, we followed the mean average precision (mAP) and the Cumulative Matching Characteristic for various ranks (rank-1 (R-1), rank-5 (R-5), rank-10 (R-10), and rank-20 (R-20)). 
We also deploy re-ranking (RR) \cite{reranking} and spatial-temporal statistics (ST)\cite{market2} for evaluation. 

\subsection{Implementation Details}
\label{section:implement}
The entire model is trained in an end-to-end fashion. For MARS and PRID-2011 datasets frame’s input dimensions are $250\times150$ with $t=4$. We set the batch size $B=32\times5$ and the number of positive instances per class $K=5$. The iLIDS-VID dataset has input size of $220\times150$ with $t=5$, $B=28\times5$, and $K=5$. For all image person ReID datasets, we set the input size to $250\times150$, with $B=32\times4$, $K=4$ and $t=1$. For the VeRi-776 dataset, the input size is $150\times250$, with the task-specific backbone CNN pre-trained on the VehicleID dataset \cite{vehicle4} instead of MARS. During the training, we include bag-of-tricks proposed by Luo \etal \cite{bot}. We average up to $32$ clip embeddings of $t$ length per identity for evaluating videos, while for images, we average all query and gallery set embeddings per identity. Our unified approach for image and video ReID is our \textbf{baseline}, with no pre-trained weights (and no pre-trained backbone CNN). Baseline with one ResNet-50 shared across branches is referred to as \textbf{baseline-1R}. The baseline with pre-trained backbone CNN is \textbf{FGReID} (Figure \ref{fig:video_architecture}). The end-to-end pre-training on MARS is denoted by \textbf{FGReID*}. 

\subsection{Comparison with State-of-the-art Methods}
\label{section:comparison}
The following section compares FGReID with existing state-of-the-art (SOTA). We report re-rank (RR) accuracy separately to maintain uniformity with previous works.
The success of previous work involving temporal self-attention (GLTR \cite{ilvid3}) and multi-granular attentive features \cite{mars_sota} supports our use of non-local block (self-attention) and fine-grained spatial attention. The inferior performance of baseline hints at the necessity of MARS pre-training. Results indicated as "-" were not reported in the published material. 

\subsubsection{Video ReID}

\begin{table}[!t]
\begin{center}
\begin{tabular}{|l|c|c|c|c| }
\hline
Method & mAP & R-1 &  R-5 &  R-20 \\
\hline\hline
\multicolumn{5}{|c|}{w/o re-ranking (RR)}\\
\hline
MGH  \cite{graph4} & 85.8 & \textbf{90.0} &  96.7  & 98.5 \\ 
SG-RAFA \cite{mars_sota} & 85.1 & 87.8 & 96.1 & 98.6 \\ 
MG-RAFA(N=2) \cite{mars_sota} & 85.5 & 88.4 & \textbf{97.1} & 98.5 \\ 
MG-RAFA(N=4) \cite{mars_sota} & 85.9 & 88.8 & 97.0 & 98.5 \\ 
\rowcolor{Gray}
Baseline-1R & 71.8 & 81.2 & 92.8 & 96.6 \\
\rowcolor{Gray}
Baseline  & 81.5 & 86.7 & 95.0 & 98.3 \\
\rowcolor{Gray}
FGReID  & \textbf{86.2} & 89.6 & 97.0 & \textbf{98.8} \\
\hline
\multicolumn{5}{|c|}{with  re-ranking (RR)}\\
\hline
Pathak \etal \cite{pathak2019video} & 88.5 & 88.0 & 96.1 & 98.5 \\
\rowcolor{Gray}
Baseline  & 88.1 & 87.5 & 95.4 & 98.5 \\
\rowcolor{Gray}
FGReID &\textbf{89.6} & \textbf{88.8} & \textbf{96.7} & \textbf{98.8}\\
\hline
\end{tabular}
\end{center}
\caption{FGReID Performance on MARS dataset.}
\label{tab:mars_performance}
\end{table}

\begin{table}[!t]
\begin{center}
\begin{tabular}{|l|c|c|c|c|}
\hline
Method   &  R-1 &  R-5  & R-10 & R-20 \\ 
\hline \hline
GLTR \cite{ilvid3} & 95.5 & \textbf{100} & \textbf{100} & \textbf{100} \\
MG-RAFA(N=4) \cite{mars_sota}  & 95.9 & 99.7 & - & \textbf{100} \\
\rowcolor{Gray}
Baseline-1R & 82.3 & 94.4 & 97.4 & 99.6 \\
\rowcolor{Gray}
Baseline  & 93.6 & 98.2  & 99.3 & 99.9  \\
\rowcolor{Gray}
FGReID & \textbf{96.1} & 99.1 & 99.9 & \textbf{100} \\
\hline
\end{tabular}
\end{center}
\caption{Results comparison for the PRID-2011 dataset.}
\label{tab:prid_performance}
\end{table}

Zhang \etal \cite{mars_sota} proposed aggregating spatial and temporal features by splitting channels into $S$ groups (SG-RAFA, $S=1$) and reading fine-grained spatial details on $N$ granular scales (MG-RAFA, $N=2,4$). Table \ref{tab:mars_performance} shows FGReID surpassing SOTA by 0.3\% on mAP and 0.2\% on R-20 accuracy without re-rank (RR) on the MARS dataset. With re-rank, FGReID beats SOTA by 1.1\% on mAP, 0.8\% on R-1, 0.6\% on R-5 and 0.3\% on R-20 accuracy. For PRID-2011, we focused on R-1 accuracy while averaging ten splits, despite having 100\% R-5 accuracy on some runs. Table \ref{tab:prid_performance} shows FGReID beating SOTA by 0.2\% on R-1 accuracy. We don't compare our results with Pathak \etal \cite{pathak2019video} as they reported their performance only for one split. On iLIDS-VID, FGReID outperforms existing SOTA significantly, with a margin of 2.9\% on R-1, and 1.2\% on R-5 accuracy, with 100\% accuracy on R-20 (Table \ref{tab:ilvid_performance}). 

\begin{table}[!t]
\begin{center}
\begin{tabular}{|l|c|c|c|}
\hline
Method   &  R-1 &  R-5  & R-20 \\ 
\hline \hline
GLTR \cite{ilvid3} & 86.0 & 98.0 & - \\
Zhao \etal \cite{ilvid2} &  86.3 & 97.4 & 99.7   \\
MG-RAFA (N=4)\cite{mars_sota} & 88.6 & 98.0 &  99.7 \\
\rowcolor{Gray}
Baseline-1R & 62.4 & 82.9 & 93.7 \\
\rowcolor{Gray}
Baseline  & 83.2 & 95.5  &  99.2  \\
\rowcolor{Gray}
FGReID & \textbf{91.5} & \textbf{99.2} & \textbf{100} \\
\hline
\end{tabular}
\end{center}
\caption{Results comparison for the iLIDS-VID dataset. FGReID achieves 99.8\% for R-10 accuracy.}
\label{tab:ilvid_performance}
\end{table}

\begin{figure*}[!t]
 \centering
 \includegraphics[width=0.9\linewidth]{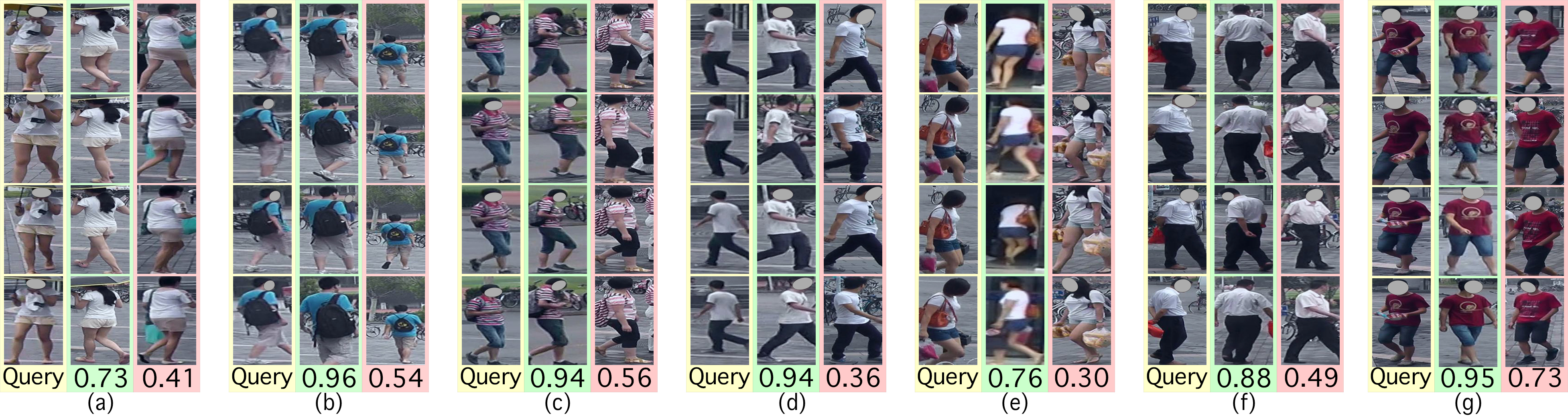}
 \caption{Yellow columns show some query tracklets, while green and red are similar-looking tracklets from the MARS dataset's gallery set. Green ones are correct R-1 results, while red ones are wrong. The scores highlight the dot product of the query with gallery embedding.}
 \label{fig:mistaken_identity}
\end{figure*}

\subsubsection{Image ReID}
\noindent For the Market1501 dataset, the original st-ReID (SOTA) \cite{market2} uses a simple part-based model with spatial-temporal statistics (ST). Table \ref{tab:market_performance} shows FGReID pre-trained on MARS dataset (with ST) surpasses SOTA by 3.6\% on mAP, 0.2\% on R-1, and 0.1\% on R-5 accuracy, w/o RR. With RR, FGReID surpasses SOTA by 0.2\% on mAP accuracy. For CUHK01, Table \ref{tab:chuk_performance} shows the performance of FGReID on both the splits, p=486 and p=100. Table \ref{tab:veri_performance} shows FGReID exceeding PRN (SOTA)\cite{veri1} on R-1 accuracy by 1.3\% (w/o RR) on the VeRi-776 dataset. 

\begin{table}[!ht]
\begin{center}
\begin{tabular}{|l|c|c|c|}
\hline
Method  & mAP &  R-1 &  R-5  \\ 
\hline \hline
\multicolumn{4}{|c|}{w/o re-ranking (RR) }\\
\hline
st-ReID (ST) \cite{market2} & 87.6 & 98.1 &  99.3 \\ 
Adaptive L2 Reg \cite{market1} & 88.9  & 95.6 & -  \\
\rowcolor{Gray}
Baseline-1R & 58.0 & 79.0 & 91.2  \\
\rowcolor{Gray}
Baseline & 76.3  & 89.6  & 96.0  \\
\rowcolor{Gray}
FGReID & 86.1 & 94.0 & 97.6 \\
\rowcolor{Gray}
FGReID* & 87.1 & 94.7 & 98.5 \\
\rowcolor{Gray}
FGReID + ST  & 91.0 & 98.2 & \textbf{99.4}  \\
\rowcolor{Gray}
FGReID* + ST & \textbf{92.5} & \textbf{98.3} & \textbf{99.4 }\\
\hline
\multicolumn{4}{|c|}{with re-ranking (RR)}\\
\hline
st-ReID (ST) \cite{market2} & 95.5 & 98.0 & 98.9 \\
st-ReID (ST) +UnityStyle \cite{market5} & 95.8 & \textbf{98.5} & \textbf{99.0} \\
\rowcolor{Gray}
Baseline & 86.9  & 90.9  & 95.1  \\
\rowcolor{Gray}
FGReID* + ST & \textbf{96.0} & 98.1 & \textbf{99.0} \\ 
\hline
\end{tabular}
\end{center}
\caption{Results comparison on Market1501 dataset. R-10 (and RR) accuracy for \textit{FGReID* + ST} is 99.4\% (99.3\%).}
\label{tab:market_performance}
\end{table}

\begin{table}[!t]
\begin{center}
\begin{tabular}{|l|c|c|c|c| }
\hline
Method & \multicolumn{2}{|c|}{p = 486} &\multicolumn{2}{|c|}{p = 100}  \\
\hline
& R-1 & R-5 & R-1 & R-10\\
\hline\hline
DSA-reID \cite{cuhk2} & 90.4 & \textbf{97.8} & - & - \\
BraidNet  \cite{cuhk3} & - & - & 93.04 & \textbf{99.97} \\ 
\rowcolor{Gray}
Baseline-1R & 45.8 & 70.0 & 82.5 & 97.0 \\
\rowcolor{Gray}
Baseline & 78.4 & 92.2 & 97.1 & 99.3 \\
\rowcolor{Gray}
FGReID & 89.6 & 96.7 & 98.9 & 99.8 \\
\rowcolor{Gray}
FGReID* & \textbf{90.9} & 97.5 & \textbf{99.1} & 99.8 \\
\hline
\end{tabular}
\end{center}
\caption{Results for CUHK01 dataset. Accuracy of FGReID* for p=486 (R-10) is 98.7\% and p=100 (R-5) is 99.8\%. }
\label{tab:chuk_performance}
\end{table}

\begin{table}[!t]
\begin{center}
\begin{tabular}{|l|c|c|c|}
\hline
Method  & mAP (RR)&  R-1 (RR) &  R-5 (RR)  \\ 
\hline \hline
PRN \cite{veri1} & \textbf{85.8}(\textbf{90.5})&97.1(\textbf{97.4})&\textbf{99.4}(\textbf{98.9}) \\ 
\rowcolor{Gray}
Baseline-1R & 39.8(45.4) & 70.5(72.2) & 83.4(79.4) \\
\rowcolor{Gray}
Baseline & 59.7(64.1) &  86.2(87.5) & 93.2(90.3)\\
\rowcolor{Gray}
FGReID & 72.3(74.9) &91.6(92.2)&96.2(94.5)\\
\rowcolor{Gray}
FGReID + ST & 84.2(86.1) & \textbf{98.4}(95.8) & 99.0(96.9) \\ 
\hline
\end{tabular}
\end{center}
\caption{Results comparison on the VeRi-776 dataset. R-10 (and RR) accuracy for the \textit{FGReID + ST} is 99.5\% (98.3\%).}
\label{tab:veri_performance}
\end{table}

\section{Ablation Study}

 \begin{table}[!t]
\begin{center}
\begin{tabular}{|l|c|c|c|c|c|}
\hline
Architectural Change & R-1 &  R-5 & R-10 & R-20 \\
\hline\hline
FGReID &  \textbf{91.5} &  \textbf{99.2} & \textbf{99.8}  & \textbf{100.0} \\
w/o $S_{channel}$  & \textbf{91.5} & 98.9 & \textbf{99.8}  & \textbf{100.0} \\
w/o non-local block &  \textbf{91.5} & 99.1 & \textbf{99.8}  & \textbf{100.0} \\
KQV  & \textbf{91.5} & 98.9 & 99.7 & \textbf{100.0} \\
\hline
Baseline-1R  & 62.4 & 82.9 & 89.5  & 93.7  \\
only GFM  & 61.4  & 82.3  & 89.3 & 94.7 \\
w/o GFM  & 87.0  & 97.6  & 99.0   & 99.7 \\
Baseline-1R (MARS) & 87.9 & 97.7 &  99.3 & 99.9 \\
only GFM (MARS) & 86.8 & 97.3  & 98.8 & 99.8  \\
w/o FGM (spatial attention) & 91.2  & 99.1  & 99.7   & 99.9 \\
\hline
\end{tabular}
\end{center}
\caption{Impact of various components (Section \ref{section:sec_components}) on FGReID’s performance,
evaluated on iLIDS-VID dataset}
\label{tab:components}
\end{table}

\subsection{Analysis of various components}
\label{section:sec_components}
Section \ref{section:methodology} and Section \ref{section:implement} describe FGReID and baseline-1R. We show the relevance of each component of FGReID by measuring the performance drop on the iLIDS-VID dataset when removing that component. As shown in Table \ref{tab:components}, the channel weights $S_{channel}$ (Equation \ref{eq:normalization}) and the non-local block have a minimal contribution in ReID. FGReID uses a variant of the non-local block, where Query (Q) and Key (K) vectors are the same. We observe using the original non-local block (KQV) of using  distinct K and Q vectors shows no performance boost. 

Baseline-1R’s poor performance owes to the single ResNet’s features multi-tasking to cover surrounding information (global features) and highlight fine-grained details. Massive performance drop in FGReID containing only the Global Feature Module (GFM) hints at the necessity of fine-grained details in ReID. Similarly, without GFM, the performance drops as FGReID solely relies on fine-grained features, indicating the need for surrounding information. MARS pre-training, indicated by (MARS), experiences a considerable performance gain, highlighting the inefficiency of ImageNet pre-training in ReID. Section \ref{section:comparison} compares Baseline-1R and Baseline with FGReID, hinting at MARS pre-training as the major reason for performance gain. The extra training parameters from extra ResNet-50 and the large size of the MARS dataset aren’t the primary reason for performance gain as the case with Market1501, where the target dataset has the same size as the source MARS dataset (Table \ref{tab:market_performance}). Without the Fine-Grained Module (FGM), the model uses spatial averaging with a slight performance drop, indicating a small contribution from spatial attention. In such a case, traditional convolutional spatial attention would add training parameters with minimal performance gain, favoring our case of parameterless spatial attention maps.

\subsection{Memory and Computation Time Comparison}
We compare the publicly available ReID models \wrt to their memory sizes, train time (forward and backward passes, processing $16$ frames per identity), and evaluation time (forward pass, processing up to $32\times4$ frames per identity) under identical conditions on MARS datasets. We train the model for $100$ epochs averaging the last 50 epochs. Similarly, we run evaluations (eval) ten times, averaging the final five runs. Table \ref{tab:computation_comparison} shows FGReID’s much faster speed compared to the multi-granular approach MGH \cite{graph4} (similarly for other such methods like MG-RAFA \cite{mars_sota}) due to our single-pass compared to their repeated passes on multi-granular scales. GLTR \cite{ilvid3} and Pathak \etal \cite{pathak2019video} are relatively simple models with far inferior performance. Substituting one of the ResNet-50 with a lightweight backbone CNN would reduce around $23.51$ million parameters. Sharing $1\times1$ conv and batch normalization layer between branches would save around $2.36$ million parameters.

\subsection{Visualization}
\begin{figure}[!t]
 \centering
 \includegraphics[width=0.9\linewidth]{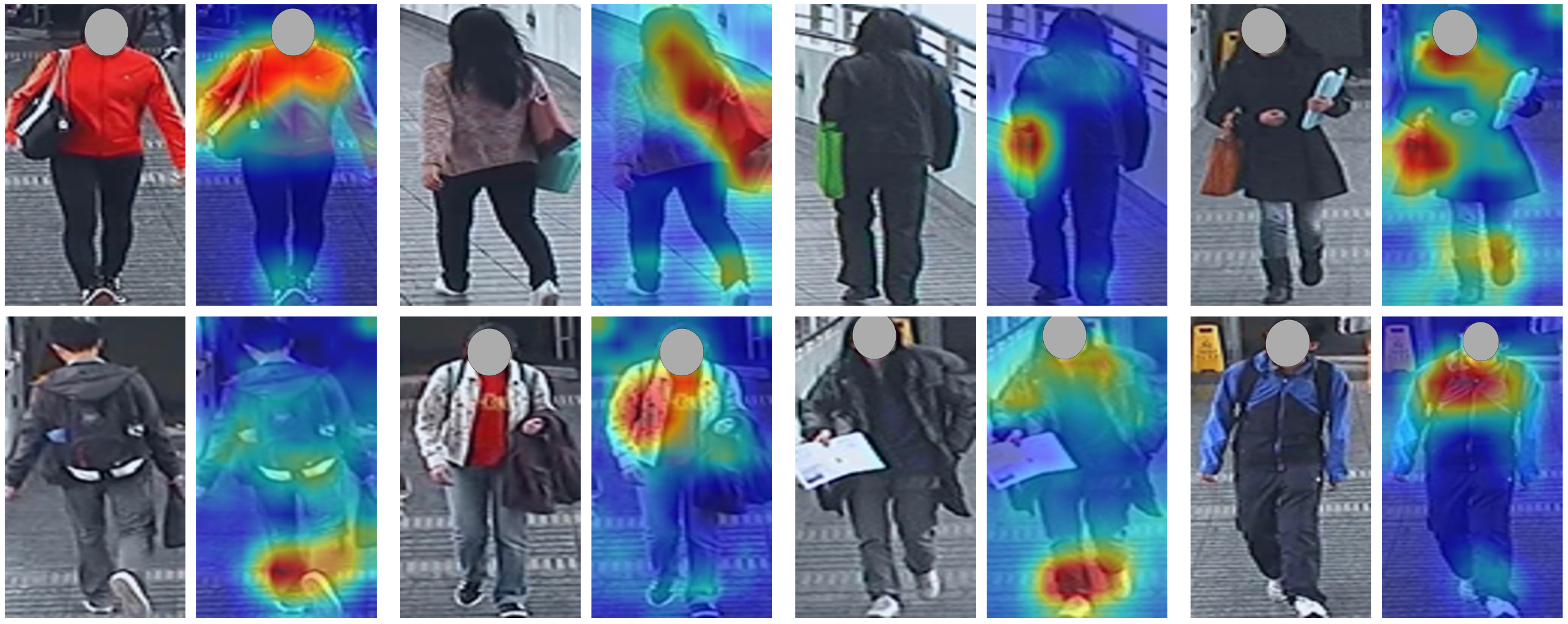}
 \caption{CUHK01 dataset samples, along with their spatial attention maps. Attention tends to focus on footwear and handbags as the primary source of fine-grained details.}
 \label{fig:attention}
 \end{figure}

We manually selected similar-looking identities (mistaken identity) from the MARS dataset to show the effectiveness of fine-grained details. Figure \ref{fig:mistaken_identity} shows a higher dot product score of the query tracklet with the correct R1 gallery tracklet compared to the mistaken identities. The scores show that the model pays attention to minute details like backpack design $\big($(a), (b), (e)$\big)$ orange handbags $\big($f$\big)$, \etc, to ReID individuals. The model is also invariant to the color structures $\big($(c), (d), (g)$\big)$. These observations are further verified by Figure \ref{fig:attention}, where spatial attention spotlights fine-grained features like handbags and shoes while disregarding background. These fine-grained regions help in differentiating subtle details in similar images.
 
\begin{table}[!t]
\begin{center}
\begin{tabular}{|l|c|c|c|}
\hline
Model & Size & Train Time  & Eval Time  \\
\hline\hline
MGH  \cite{graph4} & 44.18M & 414.90 & 719.87 \\
Pathak \etal \cite{pathak2019video} & 91.90M & 71.03 &  264.95 \\
GLTR \cite{ilvid3} & 24.77M & 62.08 & 225.43 \\ 
\rowcolor{Gray}
FGReID & 52.64M & 130.82 & 458.28 \\
\hline
\end{tabular}
\end{center}
\caption{All times are in seconds, while M means parameters count in millions.}
\label{tab:computation_comparison}
\end{table}




\subsection{Analysis of Loss functions}
To study the significance of various loss functions, we exclude one loss from the total loss during the training and assess the performance of FGReID on the iLIDS-VID dataset. Table \ref{tab:ablation_loss} shows the absence of classification loss ($L_{CE}^{avg}$),   
center loss ($L_{C}$), and variance regularization ($R_{var}$) have a significant unfavorable effect. Performance drop in the absence of KL divergence loss ($L_{cns}$), indicates a strong need for alignment of coarse-grained and fine-grained predictions. A combination of Triplet loss ($L_{trip}$) and CL Centers OSM loss ($L_{OSM}$) works best overall. We also observe the absence of satisfying rank loss ($L_{sr}$) performs better on R-1 and R-5 but has a slight drop in R-10 and R-20. 

\begin{table}[!t]
\begin{center}
\begin{tabular}{|l|c|c|c|c|c|}
\hline
$L_{total}$ & R-1 &  R-5 & R-10 & R-20 \\
\hline\hline
All Loss present & 91.5 & 99.2 & \textbf{99.8} &  \textbf{100.0} \\
w/o Cross Entropy ($L_{CE}^{avg}$) & 79.5 & 93.9 & 97.7 & 99.5\\
w/o Triplet ($L_{trip}$) & 91.1 & 98.9 & \textbf{99.8} & \textbf{100.0}\\ 
w/o OSM-CL ($L_{OSM}$) & 91.7 & 99.1 & 99.7 & \textbf{100.0} \\
w/o Center ($L_{C}$) & 88.7 & 98.1 & 99.7 & 99.9 \\
w/o Variance reg. ($R_{var}$) & 89.4 & 98.7 & 99.7 & \textbf{100.0 }\\
w/o Stratified Rank  ($L_{sr}$) & \textbf{92.1} & \textbf{99.3} & 99.6 & 99.9 \\
w/o KL-divergence ($L_{cns}$) & 88.3 & 97.7 & 99.0 & 99.7 \\
\hline
\end{tabular}
\end{center}
\caption{Impact of various loss functions on the performance of models evalauted on iLIDS-VID dataset.}
\label{tab:ablation_loss}
\end{table}

\subsection{Analysis of hyperparameters}
\begin{figure}[!t]
\centering
\subfigure[Varying K for constant P]{
\includegraphics[width=0.4\linewidth]{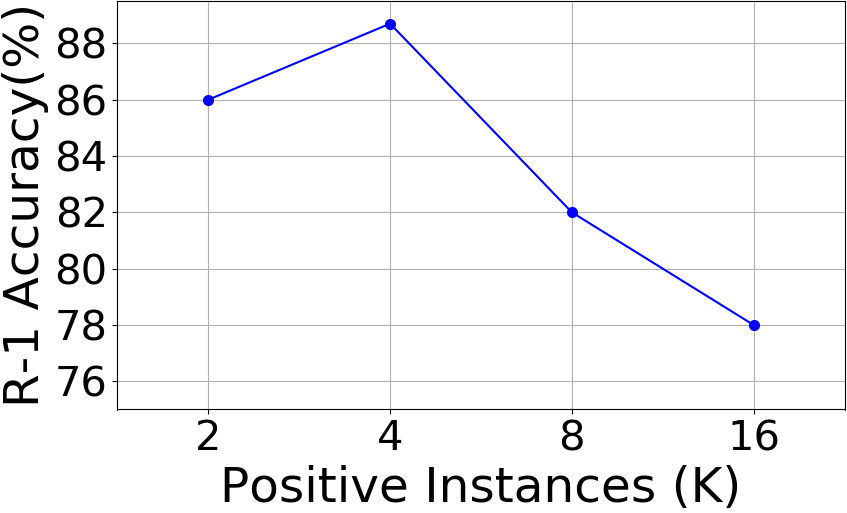}
\label{fig:32_xyz}}
\quad
\subfigure[Varying P for constant K]{
\includegraphics[width=0.4\linewidth]{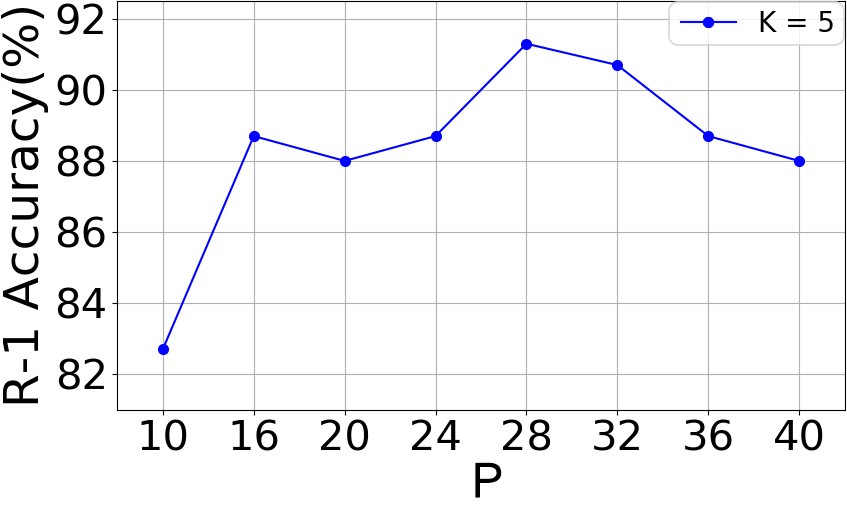}
\label{fig:xyz_5}}
\caption{(a) and (b) are evaluated on the iLIDS-VID dataset}
\end{figure}

\noindent \textbf{Batch size}
For our experiments, $B=P\times K$ and triplet loss heavily relies on the value of $P$ and $K$. We determine the optimal $P$ and $K$ for the iLIDS-VID dataset. Figure \ref{fig:32_xyz} keeps $P=32$ and varies $K$, with optimal performance at $K=4$. Figure \ref{fig:xyz_5} varies $P$ while keeping constant $K=5$, with optimal performance around $P=28$.

\begin{figure}[!t]
\centering
\subfigure[iLIDS-VID dataset]{%
\includegraphics[width=0.4\linewidth]{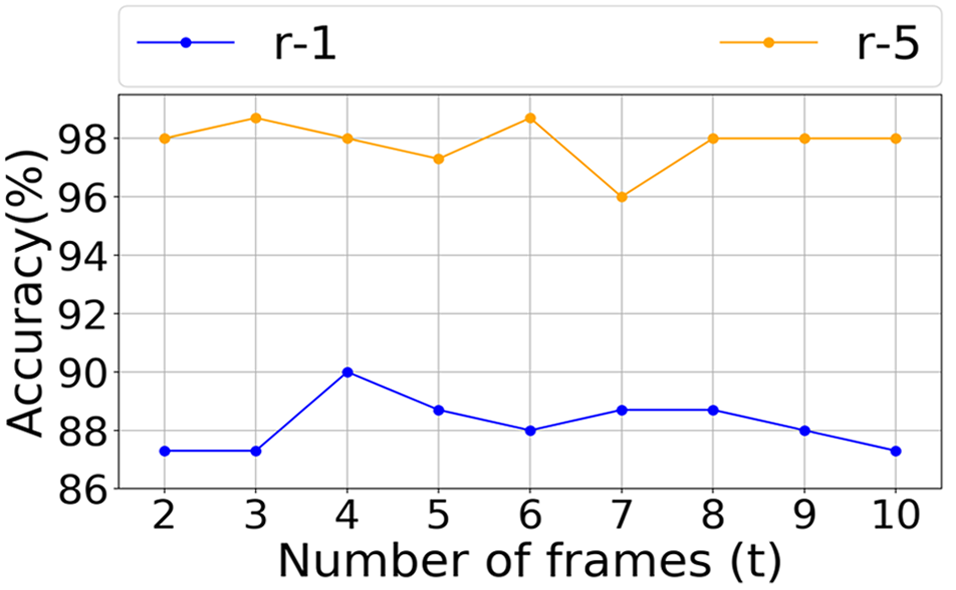}
\label{fig:seq_length}}
\quad
\subfigure[CUHK01 dataset]{%
\includegraphics[width=0.4\linewidth]{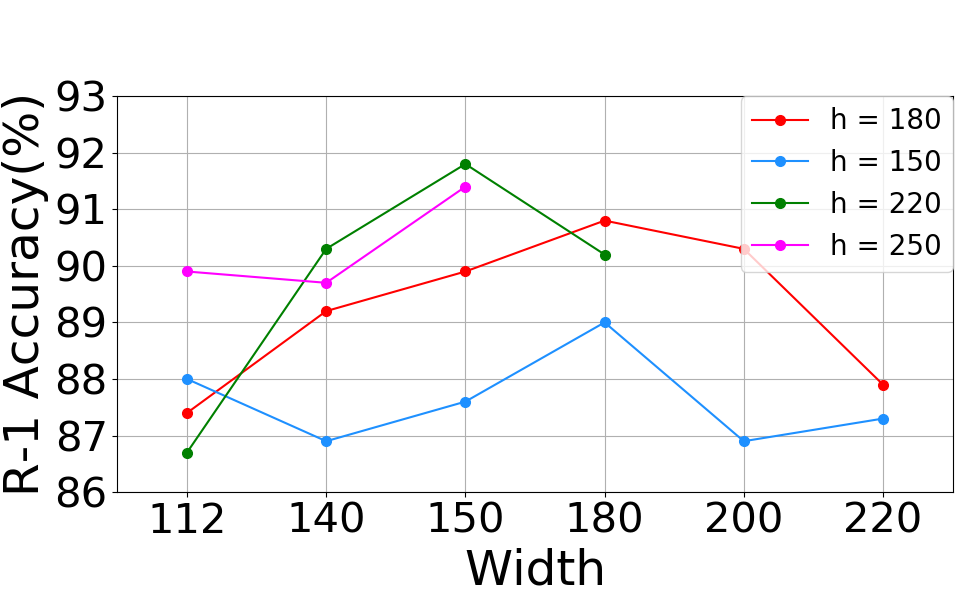}
\label{fig:width}}
\caption{(a) Compares performance for the different number of video frames (t). (b) Shows the performance impact of various input image dimensions.}
\end{figure}

\noindent \textbf{Number of Video frames}: 
For iLIDS-VID dataset, we vary the video length $t$ indicating the number of frames. Figure \ref{fig:seq_length} shows optimal performance for $t=3,4,5$. 

\noindent \textbf{Image/Frame Size}:
Most ReID works adhere to a conventional height $H$ to width $W$ ratio of 2:1, predominantly having shapes $224\times112$ and $256\times128$. We argue such a ratio may distort human proportions with Figure \ref{fig:width} showing the optimal performance, with $W=150$ and $H=220,250$. Missing regions are out-of-memory locales.

\section{Ethical Consideration}
Our approach intends to create fine-grained rich embeddings for videos and images in a zero-shot learning setting, generalizing to various embedding related tasks. Once properly deployed, ReID can spare hours of human effort in tracing suspects by reducing city-wide camera footage to a minimal subset. But possible unintended use cases exist, including the unapproved tracking of individuals and targeting protesters. The authors have genuine concerns over the alleged targeting of Uighur Muslims in China using ReID\footnote{Article: \url{https://megapixels.cc/duke_mtmc/}}. This unintentional application is undesirable, and to reduce the likelihood of it happening in the future, we have chosen not to release any hyperparameters or trained weights. All research into re-identification should consider and curtail these potentials for misuse.

\section{Conclusion}
This paper proposes FGReID, a computationally cheap method for generating contextually coherent and fine-grained rich embeddings for ReID tasks. FGReID can handle similar-looking identities and offers a unified solution for both video and image ReID tasks. The pipeline consists of three vital components:  a Global
Feature Module, a Fine-Grained Module, and a Context Module. The global feature branch delivers the general overview, while the fine-grained module highlights the minute subtleties. The context module adds temporal and spatial context via a non-local block operation to the spatially attended features. Several experiments show the viability of FGReID for both video and image ReID tasks. We perform an ablation study to highlight the importance of MARS pre-training and the need for global and fine-grained features. We opted for a single pass method of generating the above two features, which made use of an extra ResNet but reduced our training speed compared to muti-granular and partition-based approaches of generating fine-grained features. 

\section{Acknowledgement}
This work was supported in part through the NYU IT High Performance Computing resources, services, and staff expertise. Mr. Shenglong Wang (NYU), Dr.  Amir Erfan Eshratifar (USC), 
Mr. Parth Mcpherson (NYU), 
and Dr. Thomas Lux (Virginia Tech) provided their invaluable contributions to the entire project.


{\small
\bibliographystyle{ieee_fullname}
\bibliography{egbib}

\begin{thebibliography}{10}\itemsep=-1pt

\bibitem{graph1}
S. {Bai}, X. {Bai}, and Q. {Tian}.
\newblock Scalable person re-identification on supervised smoothed manifold.
\newblock In {\em 2017 IEEE Conference on Computer Vision and Pattern
  Recognition (CVPR)}, pages 3356--3365, 2017.

\bibitem{lstm1}
Xiang Bai, Mingkun Yang, Tengteng Huang, Zhiyong Dou, Rui Yu, and Yongchao Xu.
\newblock Deep-person: Learning discriminative deep features for person
  re-identification.
\newblock {\em Pattern Recognition}, 98:107036, 2020.

\bibitem{str_reid1}
L. {Bao}, B. {Ma}, H. {Chang}, and X. {Chen}.
\newblock Preserving structural relationships for person re-identification.
\newblock In {\em 2019 IEEE International Conference on Multimedia Expo
  Workshops (ICMEW)}, pages 120--125, 2019.

\bibitem{veri1}
Hao Chen, Benoit Lagadec, and Francois Bremond.
\newblock Partition and reunion: A two-branch neural network for vehicle
  re-identification.
\newblock In {\em Proceedings of the IEEE/CVF Conference on Computer Vision and
  Pattern Recognition (CVPR) Workshops}, June 2019.

\bibitem{str_reid3}
Yang Fu, Xiaoyang Wang, Yunchao Wei, and Thomas~S. Huang.
\newblock {STA:} spatial-temporal attention for large-scale video-based person
  re-identification.
\newblock {\em CoRR}, abs/1811.04129, 2018.

\bibitem{attn_better_lstm}
Jiyang Gao and Ram Nevatia.
\newblock Revisiting temporal modeling for video-based person reid.
\newblock {\em arXiv preprint arXiv:1805.02104}, 2018.

\bibitem{action_transformer}
Rohit Girdhar, Jo\~{a}o Carreira, Carl Doersch, and Andrew Zisserman.
\newblock {Video Action Transformer Network}.
\newblock In {\em CVPR}, 2019.

\bibitem{3dconv}
Xinqian Gu, Hong Chang, Bingpeng Ma, Hongkai Zhang, and Xilin Chen.
\newblock Appearance-preserving 3d convolution for video-based person
  re-identification.
\newblock In {\em ECCV}, 2020.

\bibitem{fastreid}
Lingxiao He, Xingyu Liao, Wu Liu, Xinchen Liu, Peng Cheng, and Tao Mei.
\newblock Fastreid: A pytorch toolbox for general instance re-identification,
  2020.

\bibitem{prid_d}
Martin Hirzer, Csaba Beleznai, Peter~M. Roth, and Horst Bischof.
\newblock {Person Re-Identification by Descriptive and Discriminative
  Classification}.
\newblock In {\em {Proc. Scandinavian Conference on Image Analysis (SCIA)}},
  2011.

\bibitem{fine_erase1}
Tao Hu and Honggang Qi.
\newblock See better before looking closer: Weakly supervised data augmentation
  network for fine-grained visual classification.
\newblock {\em CoRR}, abs/1901.09891, 2019.

\bibitem{variance_reg}
Ranran Huang, Hanbo Sun, Ji Liu, Lu Tian, Li Wang, Yi Shan, and Yu Wang.
\newblock Feature variance regularization: A simple way to improve the
  generalizability of neural networks, 2019.

\bibitem{Tattn2}
Tsung-Wei Huang, Jiarui Cai, Hao Yang, Hung-Min Hsu, and Jenq-Neng Hwang.
\newblock Multi-view vehicle re-identification using temporal attention model
  and metadata re-ranking.
\newblock In {\em Proceedings of the IEEE/CVF Conference on Computer Vision and
  Pattern Recognition (CVPR) Workshops}, June 2019.

\bibitem{3dconv4}
J. {Li}, S. {Zhang}, and T. {Huang}.
\newblock Multi-scale temporal cues learning for video person
  re-identification.
\newblock {\em IEEE Transactions on Image Processing}, 29:4461--4473, 2020.

\bibitem{ilvid3}
J. {Li}, S. {Zhang}, J. {Wang}, W. {Gao}, and Q. {Tian}.
\newblock Global-local temporal representations for video person
  re-identification.
\newblock In {\em 2019 IEEE/CVF International Conference on Computer Vision
  (ICCV)}, pages 3957--3966, 2019.

\bibitem{Tattn1}
Mengliu Li, Han Xu, Jinjun Wang, Wenpeng Li, and Yongli Sun.
\newblock Temporal aggregation with clip-level attention for video-based person
  re-identification.
\newblock In {\em Proceedings of the IEEE/CVF Winter Conference on Applications
  of Computer Vision (WACV)}, March 2020.

\bibitem{div_reg}
S. {Li}, S. {Bak}, P. {Carr}, and X. {Wang}.
\newblock Diversity regularized spatiotemporal attention for video-based person
  re-identification.
\newblock In {\em 2018 IEEE/CVF Conference on Computer Vision and Pattern
  Recognition}, pages 369--378, 2018.

\bibitem{cuhk_d}
Wei Li, Rui Zhao, and Xiaogang Wang.
\newblock Human reidentification with transferred metric learning.
\newblock In {\em ACCV}, 2012.

\bibitem{3dconv2}
Xingyu Liao, Lingxiao He, Zhouwang Yang, and Chi Zhang.
\newblock Video-based person re-identification via 3d convolutional networks
  and non-local attention.
\newblock In C.V. Jawahar, Hongdong Li, Greg Mori, and Konrad Schindler,
  editors, {\em Computer Vision -- ACCV 2018}, pages 620--634, Cham, 2019.
  Springer International Publishing.

\bibitem{market5}
C. {Liu}, X. {Chang}, and Y.~D. {Shen}.
\newblock Unity style transfer for person re-identification.
\newblock In {\em 2020 IEEE/CVF Conference on Computer Vision and Pattern
  Recognition (CVPR)}, pages 6886--6895, 2020.

\bibitem{vehicle4}
H. {Liu}, Y. {Tian}, Y. {Wang}, L. {Pang}, and T. {Huang}.
\newblock Deep relative distance learning: Tell the difference between similar
  vehicles.
\newblock In {\em 2016 IEEE Conference on Computer Vision and Pattern
  Recognition (CVPR)}, pages 2167--2175, 2016.

\bibitem{veri_d}
X. {Liu}, W. {Liu}, H. {Ma}, and H. {Fu}.
\newblock Large-scale vehicle re-identification in urban surveillance videos.
\newblock In {\em 2016 IEEE International Conference on Multimedia and Expo
  (ICME)}, pages 1--6, 2016.

\bibitem{vehicle1}
Xinchen Liu, Huadong Ma, Huiyuan Fu, and Mo Zhou.
\newblock Vehicle retrieval and trajectory inference in urban traffic
  surveillance scene.
\newblock In {\em Proceedings of the International Conference on Distributed
  Smart Cameras}, ICDSC '14, New York, NY, USA, 2014. Association for Computing
  Machinery.

\bibitem{lstm3}
Yiheng Liu, Zhenxun Yuan, Wengang Zhou, and Houqiang Li.
\newblock Spatial and temporal mutual promotion for video-based person
  re-identification.
\newblock {\em Proceedings of the AAAI Conference on Artificial Intelligence},
  33(01):8786--8793, Jul. 2019.

\bibitem{lstm2}
Z. {Liu}, T. {Chen}, E. {Ding}, Y. {Liu}, and W. {Yu}.
\newblock Attention-based convolutional lstm for describing video.
\newblock {\em IEEE Access}, 8:133713--133724, 2020.

\bibitem{transformer2}
Chuanchen Luo, Yuntao Chen, Naiyan Wang, and Zhaoxiang Zhang.
\newblock Spectral feature transformation for person re-identification.
\newblock In {\em Proceedings of the IEEE/CVF International Conference on
  Computer Vision (ICCV)}, October 2019.

\bibitem{bot}
Hao Luo, Youzhi Gu, Xingyu Liao, S. Lai, and W. Jiang.
\newblock Bag of tricks and a strong baseline for deep person
  re-identification.
\newblock {\em 2019 IEEE/CVF Conference on Computer Vision and Pattern
  Recognition Workshops (CVPRW)}, pages 1487--1495, 2019.

\bibitem{transformer1}
H. {Luo}, W. {Jiang}, X. {Fan}, and C. {Zhang}.
\newblock Stnreid: Deep convolutional networks with pairwise spatial
  transformer networks for partial person re-identification.
\newblock {\em IEEE Transactions on Multimedia}, pages 1--1, 2020.

\bibitem{fine_grained_5}
Dechao Meng, Liang Li, Shuhui Wang, Xingyu Gao, Zheng-Jun Zha, and Qingming
  Huang.
\newblock Fine-grained feature alignment with part perspective transformation
  for vehicle reid.
\newblock In {\em Proceedings of the 28th ACM International Conference on
  Multimedia}, MM '20, page 619–627, New York, NY, USA, 2020. Association for
  Computing Machinery.

\bibitem{market1}
Xingyang Ni, Liang Fang, and Heikki Huttunen.
\newblock Adaptivereid: Adaptive l2 regularization in person re-identification.
\newblock {\em arXiv preprint arXiv:2007.07875}, 2020.

\bibitem{pathak2019video}
Priyank Pathak, Amir~Erfan Eshratifar, and Michael Gormish.
\newblock Video person re-id: Fantastic techniques and where to find them.
\newblock In {\em Proceedings of the AAAI Conference on Artificial Intelligence
  (AAAI)}, April 2019.

\bibitem{graph2}
Yantao Shen, Hongsheng Li, Shuai Yi, Dapeng Chen, and Xiaogang Wang.
\newblock Person re-identification with deep similarity-guided graph neural
  network.
\newblock {\em CoRR}, abs/1807.09975, 2018.

\bibitem{context_frame1}
X. {Su}, X. {Qu}, Z. {Zou}, P. {Zhou}, W. {Wei}, S. {Wen}, and M. {Hu}.
\newblock k-reciprocal harmonious attention network for video-based person
  re-identification.
\newblock {\em IEEE Access}, 7:22457--22470, 2019.

\bibitem{str_reid4}
Yifan Sun, Liang Zheng, Yi Yang, Qi Tian, and Shengjin Wang.
\newblock Beyond part models: Person retrieval with refined part pooling (and a
  strong convolutional baseline).
\newblock In {\em Proceedings of the European Conference on Computer Vision
  (ECCV)}, September 2018.

\bibitem{fine_grained_3}
A. {Tan}, G. {Zhou}, and M. {He}.
\newblock Rapid fine-grained classification of butterflies based on fcm-km and
  mask r-cnn fusion.
\newblock {\em IEEE Access}, 8:124722--124733, 2020.

\bibitem{market2}
Guangcong Wang, Jianhuang Lai, Peigen Huang, and Xiaohua Xie.
\newblock Spatial-temporal person re-identification.
\newblock {\em Proceedings of the AAAI Conference on Artificial Intelligence},
  pages 8933--8940, 2019.

\bibitem{graph5}
Guan'an Wang, Shuo Yang, Huanyu Liu, Zhicheng Wang, Yang Yang, Shuliang Wang,
  Gang Yu, Erjin Zhou, and Jian Sun.
\newblock High-order information matters: Learning relation and topology for
  occluded person re-identification.
\newblock In {\em IEEE/CVF Conference on Computer Vision and Pattern
  Recognition (CVPR)}, June 2020.

\bibitem{ilvid_d}
Taiqing Wang, Shaogang Gong, Xiatian Zhu, and Shengjin Wang.
\newblock Person re-identification by video ranking.
\newblock In David Fleet, Tomas Pajdla, Bernt Schiele, and Tinne Tuytelaars,
  editors, {\em Computer Vision -- ECCV 2014}, pages 688--703, Cham, 2014.
  Springer International Publishing.

\bibitem{nonlocalattn}
Xiaolong Wang, Ross Girshick, Abhinav Gupta, and Kaiming He.
\newblock Non-local neural networks.
\newblock In {\em Proceedings of the IEEE Conference on Computer Vision and
  Pattern Recognition (CVPR)}, June 2018.

\bibitem{osm_caa}
Xinshao Wang, Yang Hua, Elyor Kodirov, Guosheng Hu, and Neil~Martin Robertson.
\newblock Deep metric learning by online soft mining and class-aware attention.
\newblock {\em CoRR}, abs/1811.01459, 2018.

\bibitem{cuhk3}
Y. {Wang}, Z. {Chen}, F. {Wu}, and G. {Wang}.
\newblock Person re-identification with cascaded pairwise convolutions.
\newblock In {\em 2018 IEEE/CVF Conference on Computer Vision and Pattern
  Recognition}, pages 1470--1478, 2018.

\bibitem{graph4}
Yichao Yan, Jie Qin, Jiaxin Chen, Li Liu, Fan Zhu, Ying Tai, and Ling Shao.
\newblock Learning multi-granular hypergraphs for video-based person
  re-identification.
\newblock In {\em Proceedings of the IEEE/CVF Conference on Computer Vision and
  Pattern Recognition (CVPR)}, June 2020.

\bibitem{graph3}
Jinrui Yang, Wei-Shi Zheng, Qize Yang, Ying-Cong Chen, and Qi Tian.
\newblock Spatial-temporal graph convolutional network for video-based person
  re-identification.
\newblock In {\em Proceedings of the IEEE/CVF Conference on Computer Vision and
  Pattern Recognition (CVPR)}, June 2020.

\bibitem{global_narrow}
Zhao Yang, Zhigang Chang, and Shibao Zheng.
\newblock Large-scale video-based person re-identification via non-local
  attention and feature erasing.
\newblock In Guangtao Zhai, Jun Zhou, Hua Yang, Ping An, and Xiaokang Yang,
  editors, {\em Digital TV and Wireless Multimedia Communication}, pages
  327--339, Singapore, 2020. Springer Singapore.

\bibitem{vehicle2}
D. {Zapletal} and A. {Herout}.
\newblock Vehicle re-identification for automatic video traffic surveillance.
\newblock In {\em 2016 IEEE Conference on Computer Vision and Pattern
  Recognition Workshops (CVPRW)}, pages 1568--1574, 2016.

\bibitem{fine_grain_erase2}
L. {Zhang}, S. {Huang}, W. {Liu}, and D. {Tao}.
\newblock Learning a mixture of granularity-specific experts for fine-grained
  categorization.
\newblock In {\em 2019 IEEE/CVF International Conference on Computer Vision
  (ICCV)}, pages 8330--8339, 2019.

\bibitem{cuhk2}
Z. {Zhang}, C. {Lan}, W. {Zeng}, and Z. {Chen}.
\newblock Densely semantically aligned person re-identification.
\newblock In {\em 2019 IEEE/CVF Conference on Computer Vision and Pattern
  Recognition (CVPR)}, pages 667--676, 2019.

\bibitem{mars_sota}
Z. {Zhang}, C. {Lan}, W. {Zeng}, and Z. {Chen}.
\newblock Multi-granularity reference-aided attentive feature aggregation for
  video-based person re-identification.
\newblock In {\em 2020 IEEE/CVF Conference on Computer Vision and Pattern
  Recognition (CVPR)}, pages 10404--10413, 2020.

\bibitem{ilvid2}
Yiru Zhao, Xu Shen, Zhongming Jin, Hongtao Lu, and Xian-sheng Hua.
\newblock Attribute-driven feature disentangling and temporal aggregation for
  video person re-identification.
\newblock In {\em Proceedings of the IEEE/CVF Conference on Computer Vision and
  Pattern Recognition (CVPR)}, June 2019.

\bibitem{mars_d}
Liang Zheng, Zhi Bie, Yifan Sun, Jingdong Wang, Chi Su, Shengjin Wang, and Qi
  Tian.
\newblock Mars: A video benchmark for large-scale person re-identification.
\newblock In Bastian Leibe, Jiri Matas, Nicu Sebe, and Max Welling, editors,
  {\em Computer Vision -- ECCV 2016}, pages 868--884, Cham, 2016. Springer
  International Publishing.

\bibitem{market_d}
Liang Zheng, Liyue Shen, Lu Tian, Shengjin Wang, Jingdong Wang, and Qi Tian.
\newblock Scalable person re-identification: A benchmark.
\newblock In {\em Proceedings of the 2015 IEEE International Conference on
  Computer Vision (ICCV)}, ICCV '15, page 1116–1124, USA, 2015. IEEE Computer
  Society.

\bibitem{reranking}
Zhun Zhong, Liang Zheng, Donglin Cao, and Shaozi Li.
\newblock Re-ranking person re-identification with k-reciprocal encoding.
\newblock {\em CoRR}, abs/1701.08398, 2017.

\bibitem{market3}
J. {Zhou}, B. {Su}, and Y. {Wu}.
\newblock Online joint multi-metric adaptation from frequent sharing-subset
  mining for person re-identification.
\newblock In {\em 2020 IEEE/CVF Conference on Computer Vision and Pattern
  Recognition (CVPR)}, pages 2906--2915, 2020.

\bibitem{fine_grained_6}
Q. {Zhou}, B. {Zhong}, X. {Lan}, G. {Sun}, Y. {Zhang}, B. {Zhang}, and R. {Ji}.
\newblock Fine-grained spatial alignment model for person re-identification
  with focal triplet loss.
\newblock {\em IEEE Transactions on Image Processing}, 29:7578--7589, 2020.

\bibitem{finegrained}
Youxiang Zhu, Ruochen Li, Yin Yang, and Ning Ye.
\newblock Learning cascade attention for fine-grained image classification.
\newblock {\em Neural Networks}, 122:174--182, 2020.

\end{thebibliography}
}

\end{document}